\documentclass[letterpaper]{article} 
\usepackage{aaai25}  
\usepackage{times}  
\usepackage{helvet}  
\usepackage{courier}  
\usepackage[hyphens]{url}  
\usepackage{graphicx} 
\usepackage{natbib}  
\usepackage{caption} 
\usepackage{algorithm}
\usepackage{algorithmic}

\usepackage{newfloat}
\usepackage{listings}
\DeclareCaptionStyle{ruled}{labelfont=normalfont,labelsep=colon,strut=off} 
\floatstyle{ruled}
\newfloat{listing}{tb}{lst}{}
\floatname{listing}{Listing}
\nocopyright 
\usepackage{newtxmath}
\usepackage{booktabs}
\usepackage{multirow}
\usepackage{subfigure}
\usepackage{float}
\usepackage{cleveref}

\title{FLIER: Few-shot Language Image Models Embedded with Latent Representations}
\author{
    Zhinuo Zhou\textsuperscript{\rm 1},
    Peng Zhou\textsuperscript{\rm 2},
    Xiaoyong Pan\textsuperscript{\rm 1}\thanks{Corresponding authors.}
}

\affiliations{
    \textsuperscript{\rm 1}Shanghai Jiao Tong University\\
    \textsuperscript{\rm 2}Second Institute of Oceanography, Ministry of Natural Resources
}

\begin{document}

\maketitle

\begin{abstract}
As the boosting development of large vision-language models like Contrastive Language-Image Pre-training (CLIP), many CLIP-like methods have shown impressive abilities on visual recognition, especially in low-data regimes scenes. However, we have noticed that most of these methods are limited to introducing new modifications on text and image encoder. Recently, latent diffusion models (LDMs) have shown good ability on image generation. The potent capabilities of LDMs direct our focus towards the latent representations sampled by UNet. Inspired by the conjecture in CoOp that learned prompts encode meanings beyond the existing vocabulary, we assume that, for deep models, the latent representations are concise and accurate understanding of images, in which high-frequency, imperceptible details are abstracted away. In this paper, we propose a Few-shot Language Image model Embedded with latent Representations (FLIER) for image recognition by introducing a latent encoder jointly trained with CLIP’s image encoder, it incorporates pre-trained vision-language knowledge of CLIP and the latent representations from Stable Diffusion. We first generate images and corresponding latent representations via Stable Diffusion with the textual inputs from GPT-3. With latent representations as “models-understandable pixels”, we introduce a flexible convolutional neural network with two convolutional layers to be the latent encoder, which is simpler than most encoders in vision-language models. The latent encoder is jointly trained with CLIP’s image encoder, transferring pre-trained knowledge to downstream tasks better. Experiments and extensive ablation studies on various visual classification tasks demonstrate that FLIER performs state-of-the-art on 11 datasets for most few-shot classification.
\end{abstract}

%

\section{Introduction}

Numerous efficient models for addressing visual tasks (such as ResNet\cite{he2016deep}, ViT\cite{dosovitskiy2020image}, CLIP\cite{radford2021learning}, BEiT\cite{bao2021beit}, MAE\cite{he2022masked}, CAE\cite{chen2024context}), have achieved continuous improvements in model performance across various datasets. However, it is observed that in real-world application scenarios, the quantity or quality of available data often falls significantly below the standards of existing comprehensive datasets. CLIP, as a representative model of visual-language fusion, holds significant importance in achieving strong few-shot generalization performance across various computer vision tasks, particularly for zero-shot image classification. Since the introduction of CLIP, numerous models have emerged aiming to optimize its performance, including CoOp\cite{zhou2022learning}, Clip-Adapter\cite{gao2023clip}, Tip-Adapter\cite{zhang2022tip}, CaFo\cite{zhang2023prompt}. Of these models, some optimize the performance of CLIP through prompt learning. Others enhance the model's ability through adaptation and cache methods.

Recently, generative models have rapidly been developed. Initially, when generative models were first introduced, a series of models, like Generative Adversarial Networks (GAN)\cite{goodfellow2014generative} and variational autoencoders (VAE)\cite{kingma2014auto}, emerged prominently. GANs operate through a discriminator and a generator to accomplish the process of image generation, generating high resolution images with good quality. Different from GANs, VAE generates high resolution images efficiently through stochastic variational inference and learning. After GANs and VAE, diffusion models, like DALL-E\cite{ramesh2021zero} and Stable Diffusion 2\cite{rombach2022high}, have gained considerable attention due to their superior generation results. By introducing noise to degrade images, diffusion models employ a learned reverse process, effectively denoising to reconstruct the original images based on stochastic processes. In the generation process of Stable Diffusion 2 (SD2), it utilizes CLIP to encode the condition information and employs a trained sampler to sample Gaussian noise, obtaining image embeddings in the latent space. Finally, embeddings are decoded through a decoder to generate images. We notice that when image embeddings are transformed into images, they may be unclear for humans to recognize. However, the decoder can decode these embeddings and produce accurate images. This observation leads us to pose a question: "Considering these embeddings are created by models themselves, are they potentially more understandable for models as latent representations?". With this in mind, we integrate the image embeddings into CLIP (the CLIP image encoder to be more specific) via simple Convolutional neural networks (CNN)\cite{krizhevsky2012imagenet}, aiming to enhance the visual-language model.

\begin{figure*}[t]
\centering
\includegraphics[height=7cm]{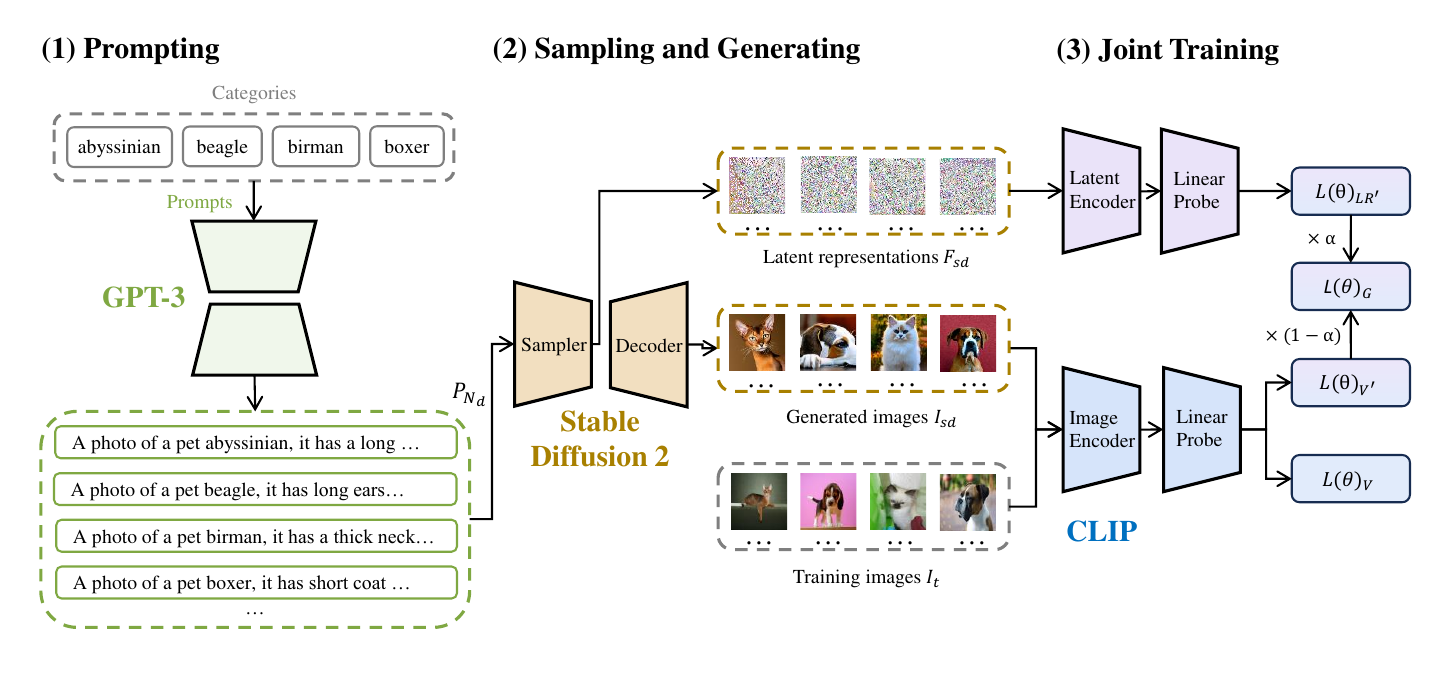}
\caption{Overview of FLIER. First, we use GPT-3 to generate prompts for each class. Then, we input the prompts into SD2 and obtain latent representations and generated images. Finally, the latent encoder and CLIP's image encoder are jointly trained with latent representations, generated images $I_{N_d, K'}$ and training images $I_{N_d, K}$.}
\label{fig:1}
\end{figure*}

In this paper, we propose FLIER, a novel visual-language model integrated with generative modules, for few-shot image recognition. In \Cref{fig:1}, the pipeline of FLIER starts from Prompting. To generate input textual prompts for SD2 corresponding to given category names, we employ GPT-3 to get rich additional information relevant to each category. Then we utilize SD2 to generate images from texts in Prompting. In the process of conditioned generation of SD2, we save the latent embeddings along with their corresponding generated images by the decoder, which not only enlarges the few-shot training data, but also gets the latent representations. Finally for joint training, we divide the training into two parts. In the first part, the image encoder of CLIP is trained by connecting linear probe on the training images from the original dataset. In the second part, the input of the image encoder is the generated images. Meanwhile, the latent encoder, a simple CNN with two convolutional layers, takes the latent representations as input. Finally, the models are trained jointly, and the loss is calculated by the weighted losses from the two models with a latent factor $\alpha$. This approach allows CLIP to be thoroughly trained on the original dataset, while incorporating latent representations into the training process in a straightforward way.

Our main contributions are summarized as follows:

1.  We propose FLIER, a new visual-language model based on CLIP for few-shot classification tasks, to integrate prior vision-language knowledge with latent representations in diffusion modules for better representation learning.

2.  Different from using frozen pre-trained weights, FLIER trains the image encoder of CLIP with the latent encoder jointly, which guarantees a new understanding of images injected into CLIP, improving FLIER’s ability on transferring prior pre-trained knowledge and latent representations to downstream classification tasks with a state-of-the-art (SOTA) on ImageNet\cite{deng2009imagenet}.

3.	We perform extensive ablation studies of FLIER on ImageNet to show the effectiveness of individual modules and evaluate FLIER on 11 benchmark datasets for few-shot classification, where FLIER achieves SOTA in most experiments without using additional annotated data.

\section{Related Work}
\subsection{Vision-Language Models}
With the continuous development of language and vision models, there has been an increasing number of methods for exploring the interaction between vision and language. After the effectiveness of attention mechanisms was proved, in vision-language models, attention-based approaches have demonstrated excellent performance such as BAN\cite{kim2018bilinear}, Intra-Inter\cite{gao2019dynamic} and MCAN\cite{yu2019deep}. Since BERT\cite{devlin2018bert} showed impressive performance, several subsequent works\cite{lu2019vilbert,tan2019lxmert} based on BERT further propelled the development of visual language models.

Recent vision-language models\cite{furst2022cloob,jia2021scaling,li2021supervision}, represented by CLIP\cite{radford2021learning}, connect visual and language knowledge by learning image and text encoders jointly. Compared to previous models, these models leverage contrastive representation learning\cite{chen2020simple,he2020momentum,henaff2020data} to fully utilize the general knowledge learned from large-scale training datasets. The remarkable success of CLIP in zero-shot image recognition inspires related research on CLIP-like models\cite{dong2022clip,zhou2022learning,zhou2022conditional,gao2023clip,zhang2022tip}. Some CLIP-based methods like CoOp\cite{zhou2022learning} introduce prompt design to improve models. Adaption\cite{gao2023clip,zhang2022tip} and fine-tuning\cite{dong2022clip} on CLIP also achieve outstanding performance. As the development of image generation, diffusion models attract great attention. Based on both vision-language models and generative models, CaFo\cite{zhang2023prompt} incorporates the knowledge of CLIP\cite{radford2021learning}, DINO\cite{caron2021emerging}, DALL-E\cite{ramesh2021zero} and GPT-3\cite{brown2020language} for improving the few-shot classification performance. Different from CaFo, we introduce a new idea about "models-understandable pixels" and our FLIER jointly trains a latent encoder for latent representations with CLIP's image encoder. Specifically, the latent encoder is a simple backbone consisting of two convolutional layers.

\subsection{Diffusion Models}
Recently, in the field of density estimation \cite{kingma2021variational} and sample quality \cite{dhariwal2021diffusion}, compared to previous methods, diffusion probabilistic models (DM) \cite{sohl2015deep} demonstrate excellent performance. These models use UNet as the backbone for generating images to fit inductive deviations naturally\cite{dhariwal2021diffusion,ho2020denoising,ronneberger2015u,song2020score}. However, due to the complicated model structure and operating pixel space, evaluating and training these models has shortcomings that they infer slowly and cost a lot when training. To address these drawbacks, SD2\cite{rombach2022high} introduces cross-attention layers into the architecture and changes the operating space to lower dimensional compressed latent space in a two-stage image synthesis way. It speeds up the inference with almost no reduction in synthesis quality. Inspired by the exciting results of SD2 and the detachable architecture, we isolated the latent representations generated by UNet during the inference for subsequent joint training with CLIP.

\subsection{Few-Shot Learning}
Few-shot learning (FSL) aims to enable models to learn and generalize to classes with a limited number of annotated samples\cite{wang2020generalizing}. In recent years, FSL typically focuses on three perspectives: data, model, and algorithm, to improve model’s performance. At the data level, FSL often employs prior knowledge to augment and enhance datasets\cite{douze2018low,pfister2014domain}. In the meantime, similar datasets can provide relative prior knowledge\cite{gao2018low,tsai2017improving}. At the model level, FSL focuses on narrowing the hypothesis space and searching optimal parameters for the model with prior knowledge. Some model-based optimization methods\cite{benaim2018one,liu2018learning,cai2018memory,ramalho2019adaptive} primarily enhance the structure and design of models through four aspects: multitask learning, embedding learning, learning with external memory, and generative modeling. Meanwhile, algorithmic optimization methods\cite{kozerawski2018clear,yu2018diverse,rusu2018meta,ravi2016optimization} have traditionally concentrated on improving search strategies: how to refine existing parameters, refine meta-learning parameters, and learn optimizers.

With the emergence of visual-language pre-training models such as CLIP\cite{radford2021learning}, optimizing vision-language models on few-shot datasets has attracted considerable attention. Some works\cite{zhou2022learning,zhang2023prompt} do not require additional training of the model; instead, they achieve good performance on few-shot datasets by optimizing prompts or constructing key-value cache models. Other works\cite{dong2022clip,zhang2022tip,gao2023clip} achieve higher few-shot accuracy by introducing learnable parameters for training the model. Different from existing methods, we integrate latent representations with CLIP using simple customized networks.

\section{FLIER Approach}
An overview of our approach FLIER is shown in \Cref{fig:1}. We first introduce the method of fine-tuning CLIP’s image encoder. Then, we present the details of low-dimensional latent representation, which is the key component of FLIER. Finally, we elaborate the joint training details of FLIER.

\subsection{Contrastive Language-Image Pre-training and Fine-tuning}

\subsubsection{Contrastive Language-Image Pre-training} is a vision-language model consisting of two encoders for texts and images\cite{radford2021learning}. With a backbone of a Transformer\cite{vaswani2017attention}, the text encoder produces text representations from encoded tokens of textual inputs. The backbone of image encoder is a ViT\cite{dosovitskiy2020image} or ResNet\cite{he2016deep}, which produces the feature vectors for images. When training, CLIP utilizes cosine similarity to measure the matching degree of different text-image pairs. The optimization goal is to maximizes the cosine similarities of one image with the matched text, and minimizes that with unmatched texts for every pair. For the loss function, CLIP calculates the average cross entropy loss of two encoders. We let $ I\in \mathbb{R}^{H\times W\times 3 }$ be the image input, where $H$ and $W$ is the image's height and width. With the image encoder, an image feature $ f\in \mathbb{R}^D $ is obtained from $I$, where $D$ stands for the dimension of image's feature.

\subsubsection{CLIP Fine-tuning} requires CLIP to use a linear probe evaluation protocol for downstream classification tasks. In this way, let $W \in \mathbb{R}^{D\times K}$ represents the weight of the linear layer connected to CLIP's image encoder, where $K$ represents the number of categories. The image feature $f$ is calculated through the linear layer to obtain a logit. The softmax function then converts the logits obtained by the linear layer into predicted probability $p\in \mathbb{R}^K$.

\begin{equation}
  f = ImageEncoder(I), \ \ \ logits_i = W_i^Tf,
  \label{eq:1}
\end{equation} 

\begin{equation}
  p_i = \frac{exp(logits_i)/\gamma}{\sum^N_{j=1}exp(logits_j)/\gamma}.
  \label{eq:2}
\end{equation} 
where $\gamma$ represents the temperature of Softmax, $W_i$ stands for the prototype weight vector for class $i$, and $p_i$ denotes the probability of category $i$.

When fine-tuning CLIP (with linear probe), specifically the image encoder of CLIP, we apply strategies of the simultaneously updated exponential moving average (EMA) approach, the layer-wise learning rate (LLRD) strategy and initialing learning rate in fine-tuning. We finetune the CLIP on 11 datasets for various experiments for CLIP ViT-Base/16 with $224\times224$ input resolution on the whole ImageNet with configuration, including AdamW optimizer, cosine decay learning schedule, batch size 64, learning rate 0.0001, epochs 40, weight decay 0.05 and a latent factor 0.5 .

\subsection{Prompting and Sampling Latent Representations}

\subsubsection{Prompting via GPT-3.} Considering the experimental setting of few-shot classification, the number of generated images has to be larger than $K$ when we conduct $N$-way and $K$-shot few-shot experiments. Thus, we use 10 effective prompts for each category. For different datasets $d$, let $N_d$ represent the number of categories in $d$. For every $N_d$ category, we try practical commands for GPT-3\cite{brown2020language} such as “How do you describe a [CLASS]?”, “What does a [CLASS] look like?” and “Tell me what is a [CLASS]”. Let $P_{N_d}$ stand for prompts generated by GPT-3.

\begin{equation}
  P_{N_d}=\text{GPT-3}(\text{Commands}).
  \label{eq:3}
\end{equation} 

\subsubsection{Sampling Latent Representations via SD2.} By text-to-image method of SD2, we adopt the prompts $P_{N_d}$ as input. For model strategies, we load the checkpoint of SD2 with EMA for 512$\times$512 images and use DPM\cite{lu2022dpm} sampler for the inference. Under the setting of few-shot experiments, the maximum number required of each category $N_d$ is 16. We denote the generated images of $N_d$ as $I_{N_d, K'}$ and training images of $N_d$ as $I_{N_d, K}$, where $K$ and $K'$ represent the K-shot setting in few-shot learning. In order to avoid images insufficiency, for each category, we set the batch size to be 2 and generate 20 images by 10 given prompts $P_{N_d}$, formulated as

\begin{equation}
  I_{N_d, K'}=\text{Stable Diffusion 2}(P_{N_d}).
  \label{eq:4}
\end{equation} 

For each category, we obtain 20 generated images and their corresponding 20 latent representations. Also, we denotes $F_{N_d, n, K}$ as the latent representation of $nth$ generated image in the category $K$ of dataset $N_d$. In the few-shot experiments, to maintain the low-data rules, we keep $K$ equal to $K'$ all the time.

\subsection{Joint Training.}
We present a joint training framework by embedding low-dimensional latent representations into language-image encoder with a two-layer CNN for achieving better performance. It is difficult for CLIP’s image encoder itself to fine-tune under the few-shot setting, since a small amount of data can hardly guarantee a large model to perform well in few-shot experiments. Different from appending additional learnable bottleneck of linear layers in CLIP-Adapter\cite{gao2023clip}, we train a latent encoder with CLIP’s image encoder jointly. Let $\Psi_{l}$ represent the latent encoder, $\Psi_{v}$ represent the image encoder of CLIP, $W_l$ represent the linear probe connected with $\Psi_{l}$ and $W_v$ represent the linear probe connected with $\Psi_{v}$. We obtain the embeddings by two encoders and calculate the logits by linear probe:

\begin{equation}
    \begin{aligned}
      &f_{I}=\Psi_{v}(I_{N_d, K}),\ Logits_{V}=W_v^Tf_{I}, \\
      &f_{I'}=\Psi_{v}(I_{N_d, K'}), \ Logits_{V'}=W_v^Tf_{I'}, \\
      &f_{L'}=\Psi_{l}(F_{N_d, n, K'}), \ Logits_{L'}=W_l^Tf_{L'}.
  \label{eq:5}
  \end{aligned}
\end{equation} 

  

  

Then, we adopt \Cref{eq:2} to calculate $P_{V'}=\{p_{v',i}\}_{i=1}^K$, $P_{V}=\{p_{v,i}\}_{i=1}^K$ and $P_{LR'}=\{p_{lr',i}\}_{i=1}^K$, which denotes the category vector of the image encoder on the training data, the image encoder on the generative data and latent encoder with the latent representations.

Finally, for the loss function, we utilize label smooth cross entropy loss to prevent model over-fitting during fine-tuning. For joint training, we divide the training period into two parts. The first part is for training on the training dataset, optimizing the weights of the image encoder and its linear probe with the loss $L(\theta)_{V}$, which is calculated by \Cref{eq:9} and \Cref{eq:10}. The second part is jointly training CLIP's image encoder with the latent encoder on the generated images, optimizing the weights of both encoders and their linear probe with the loss $L(\theta)_{G}$. The loss $L(\theta)_{G}$ consists of $L(\theta)_{V'}$ and $L(\theta)_{LG'}$, which are computed based on the image encoder and latent encoder, respectively. We utilize the latent factor $\alpha$ to measure the contribution of different losses to $L(\theta)_{G}$.

\begin{equation}
  y_i=
  \begin{cases}
      \frac{\epsilon}{n} & i \not =target \\
      1-\epsilon+\frac{\epsilon}{n} & i =target
  \end{cases},
  \label{eq:9}
\end{equation} 

\begin{equation}
    L(\theta)_{S}=-\frac{1}{N}\sum_i^Ny_ilog(p_{S,i}),
    \label{eq:10}
\end{equation} 


where $S$ refers to $LR'$, $V'$ and $V$; $n$ is the total number of $N_d$'s categories; $N$ is the total number of images in the dataset; $target$ is the ground-truth label; $\theta$ represents all learnable parameters; $\epsilon$ is the smoothing factor.

\section{Experiments}
\subsection{Training Settings}
\subsubsection{Datasets.}
For few-shot classification and domain generalization evaluation, we conduct experiments for FLIER on 11 image classification datasets: ImageNet\cite{deng2009imagenet}, OxfordPets\cite{parkhi2012cats}, Caltech101\cite{fei2004learning}, SUN397\cite{xiao2010sun}, Food101\cite{bossard2014food}, DTD\cite{cimpoi2014describing}, Flowers102\cite{nilsback2008automated}, EuroSAT\cite{helber2019eurosat}, UCF101\cite{soomro2012ucf101}, StanfordCars\cite{krause20133d} and FGVCAircraft\cite{maji2013fine}. Specifically, we train our FLIER with 1, 2, 4, 8, 16 shots for few-shot classification, and we also conduct fine-tuning on the full training set for comprehensive analysis. The evaluation is on the test set of each dataset.

\subsubsection{Baselines.}
For GPT-3's prompting, we utilize three different command templates for textual prompts. For each command, GPT-3 conducts 5 prompts. Then, we screened out prompts with a word count greater than 10 and pick 10 prompts for each category randomly. For SD2, we adopt DPM as the sampler and UNet as the backbone of the forward process. Additionally, we set the batch size as 2, time step in the forward process as 50 and downsampling factor as 8. With the above settings, we sample and generate low-dimensional latent representations and images via SD2. For the implementation of CLIP, when comparing to fine-tuning CLIP directly on the whole ImageNet, we utilize ViT-B/16\cite{dosovitskiy2020image} as the backbone. When conducting few-shot experiments, we adopt ResNet50\cite{he2016deep} (RN50) as the backbone of the image encoder and its aligned transformer\cite{vaswani2017attention} as the textual encoder. For fine-tuning strategies, we adopt EMA in evaluation with its momentum factor of 0.9998 and preserve model's original accuracy as well. We report the better one of two results. Also, we apply scaling, random crop, rotation and color jitter to the image for data augmentation. In the LLRD, we pick the base learning rate from 0.00005 to 0.0001 and the default increase factor of LLDR is 0.7. We adopt AdamW optimizer with the initial learning rate of 0.0001 using a cosine scheduler. The initial learning rate is modified slightly in the experiments. When training, we normally use the batch size of 64 for 40 epochs, but due to the few-shot setting and the learning rate scheduler, we need to adjust the batch size smaller in several experiments. And for few-shot experiments like 1-shot and 2-shot, we set the number of epochs larger to make model converge. We conduct all experiments with two NVIDIA GeForce RTX 3090 GPUs.

\subsection{Performance of FLIER with baseline methods}
\subsubsection{Performance on ImageNet.}
To evaluate the vision classification ability of FLIER, We compare it with CLIP-finetune which fine-tunes CLIP\cite{dong2022clip} on ImageNet's training set. We train the models on full training set of ImageNet dataset to evaluate FLIER’s performance on general sufficient data scene. As shown in \Cref{tab:1}, FLIER achieves the better performance than CLIP-finetune with the accuracy of 87.08\%, surpassing that by 1.41\%, respectively. The results show that FLIER could learn better visual representation with the help of latent representations sampled by SD2.

\begin{table}[t]
  \caption{Comparison of FLIER with the backbone of ViT-B/16 and other methods on ImageNet for fine-tuning on the full training set.
  }
  \label{tab:1}
  \centering
  \begin{tabular}{@{}l|cccccc@{}}
    \hline\rule{0pt}{8pt}
    Models  & FD-CLIP & CLIP-finetune & \textbf{FLIER}\\
    \hline\rule{0pt}{8pt}
    Accuracy  & 84.94 & 85.67 & \textbf{87.08} \\ 
    \hline
  \end{tabular}
\end{table}

\begin{figure}[t]
  \centering
  \begin{subfigure}
    {\includegraphics[width=3.9cm]{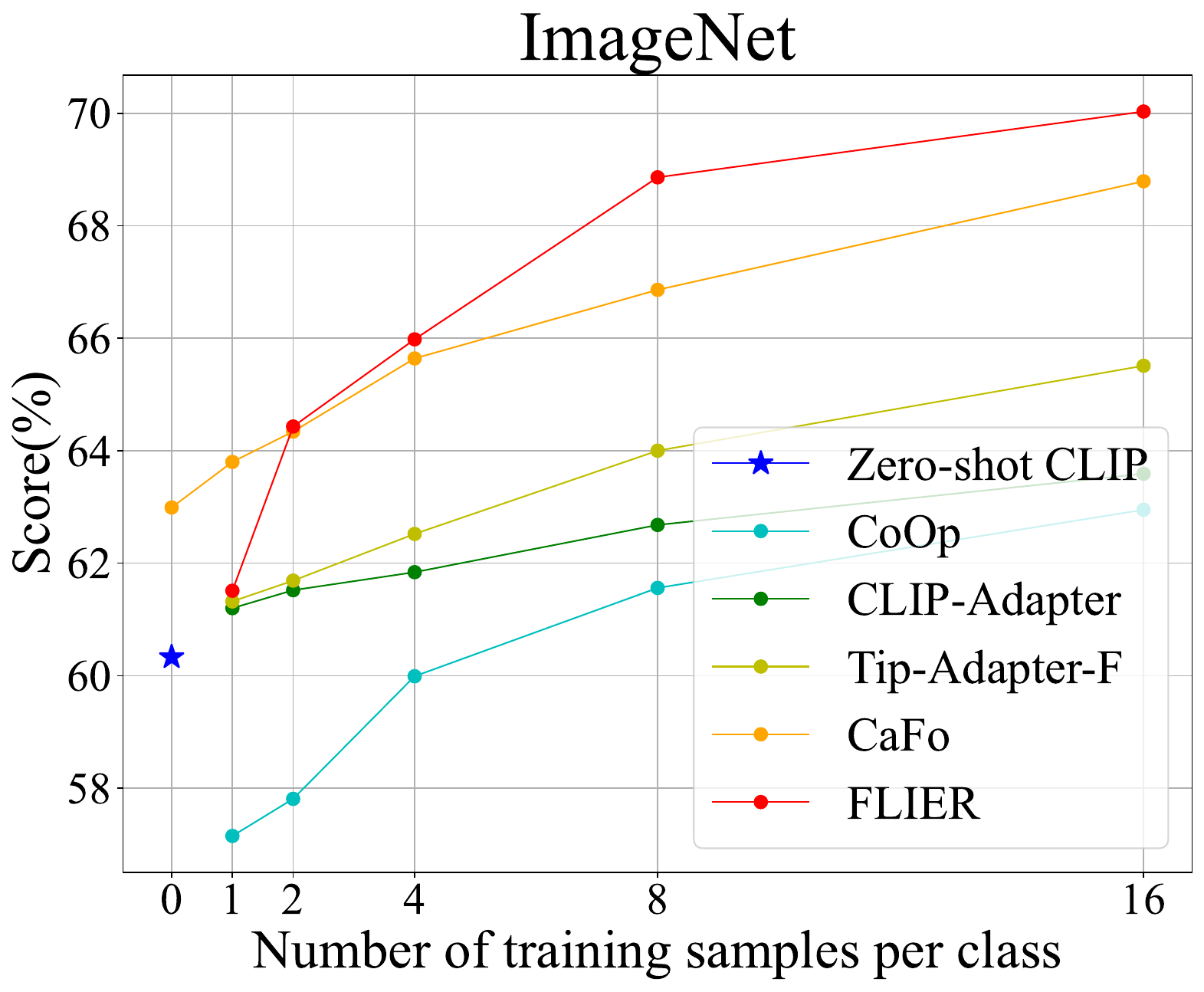} \label{fig:2-1}}
  \end{subfigure}
  \begin{subfigure}
     {\includegraphics[width=3.9cm]{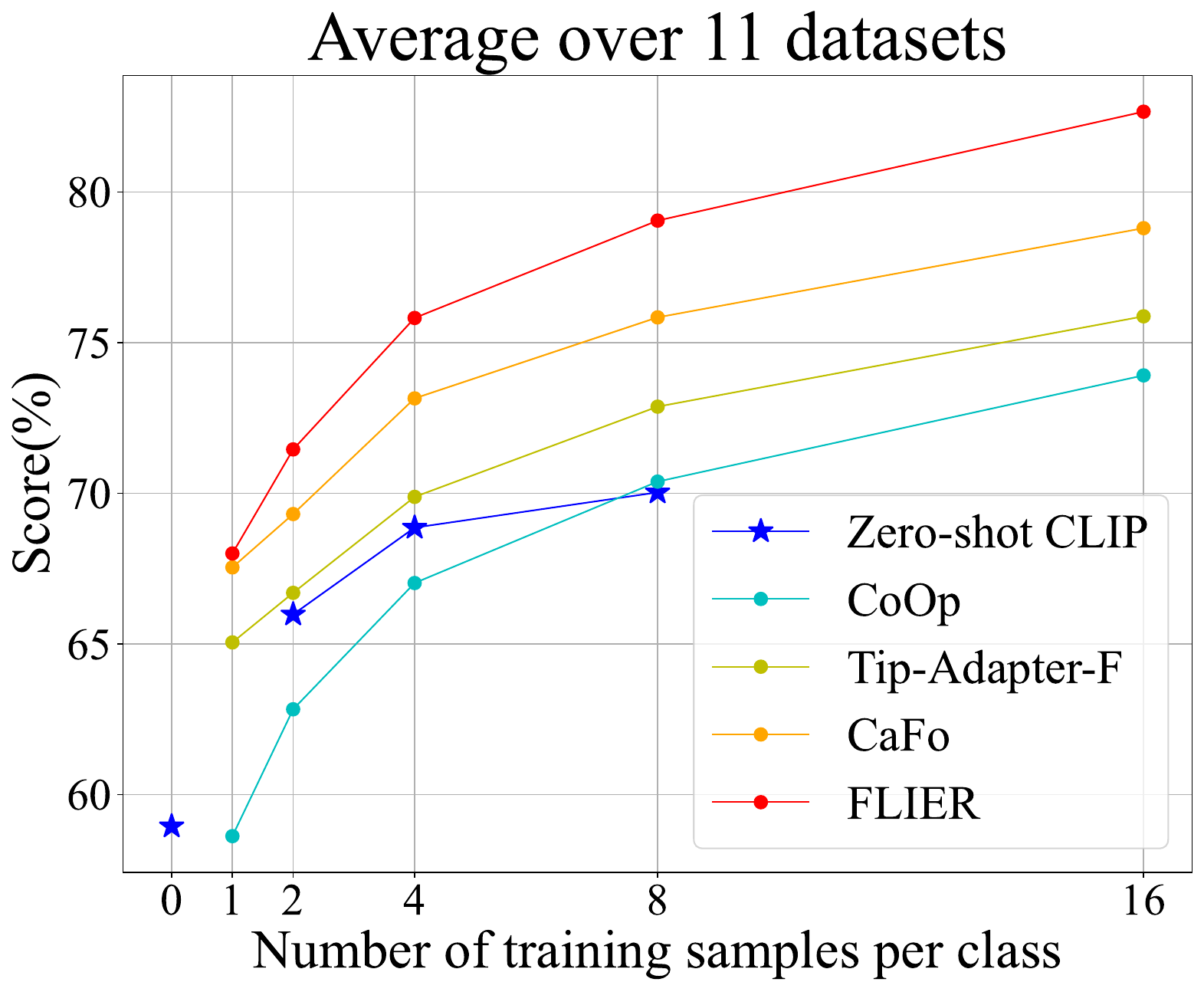} \label{fig:2-2}}
  \end{subfigure}
  \caption{Average performance results on 11 datasets and performance on ImageNet of few-shot learning. FLIER with RN50 outperforms SOTA for 2, 4, 8 and 16 shot settings on ImageNet and for average accuracy on 11 datasets.
  }
  \label{fig:2}
\end{figure}

For few-shot classification on ImageNet, we compare FLIER with other previous CLIP-based methods, including CLIP\cite{radford2021learning}, CoOp\cite{zhou2022learning}, CLIP-Adapter\cite{gao2023clip}, Tip-Adapter-F\cite{zhang2022tip} and CaFo\cite{zhang2023prompt}. We adopt pre-trained CLIP image encoders based on RN50 backbone. The results in \Cref{fig:2-1} show that FLIER performs surprisingly well on ImageNet in few-shot experiments, especially 4-shot, 8-shot and 16-shot. It also shows some advantages compared to the other methods in 1-shot and 2-shot setting. In \Cref{tab:2}, FLIER achieves the highest accuracy on the test set with (70.03\%, 68.86\%) under 16-shot and 8-shot setting, surpassing the other baseline models by (1.24\%, 2.00\%), (4.52\%, 4.86\%), (6.44\%, 6.18\%), (7.08\%, 7.30\%) and (13.90\%, 19.34\%) respectively. It is remarkable that FLIER in 8-shot setting achieves even higher accuracy (68.86\%) than previous SOTA in 16-shot setting (68.79\%). Under 1-shot and 2-shot setting, FLIER is still superior to other methods: Linear-probe CLIP (Lp-CLIP), CoOp, CLIP-Adapter and Tip-Adapter-F. But CaFo yields a competitive performance in 1-shot setting. Due to the large data required for fine-tuning, FLIER is relatively strong when the number of shots is larger or equal to 2. In addition, we analyze the efficiency of CaFo and FLIER in ViT-B/16 for their training time and accuracy. The results in \Cref{tab:3} shows that FLIER achieves a higher accuracy 76.70\% with 156 minutes of training on a single NVIDIA GeForce RTX 3090 GPU for 40 epochs, each epoch takes almost the same time as CaFo. Also, we compare the flops and parameters of FLIER with CLIP,. The results in \Cref{tab:3-1} shows that FLIER only add 0.77M parameters in the architecture to CLIP's image encoder, indicating that FLIER achieves excellent performance with a small increase of parameters.
\begin{table}[t]
  \caption{Comparison of FLIER and other methods with RN50 on ImageNet under few-shot setting.
  }
  \label{tab:2}
  \centering
  \begin{tabular}{@{}lcccccc@{}}
    \toprule
    Shot  & 1& 2 &4 & 8 & 16 \\
    \midrule
    Lp-CLIP  & 22.17 & 31.90 & 41.20 & 49.52 & 56.13 \\
    CoOp  & 57.15 & 57.81 & 59.99 & 61.56 & 62.95 \\
    CLIP-Adapter  & 61.20 & 61.52 & 61.84 & 62.68 & 63.59 \\
    Tip-Adapter-F  & 61.32 & 61.69 & 62.52 & 64.00 & 65.51 \\
    CaFo  & \textbf{63.80} & 64.34 & 65.64 & 66.86 & 68.79\\
  \bottomrule
  \textbf{FLIER}  &  61.51 & \textbf{64.43} & \textbf{65.98} & \textbf{68.86} & \textbf{70.03}
  \end{tabular}
\end{table}
\begin{table}[tb]
  \caption{Comparison of time efficiency and accuracy for FLIER and CaFo on ImageNet under 16-shot setting with a single NVIDIA GeForce RTX 3090 GPU with ViT-B/16.
  }
  \label{tab:3}
  \centering
  \begin{tabular}{@{}lcccc@{}}
    \toprule
    Models & Epoch & Time & Accuracy & Gain \\
    \midrule
    Zero-shot CLIP& 0 &0 &60.33 & - \\
    \midrule
    CaFo & \textbf{20} & \textbf{1h38min} & 74.48 & +14.15 \\
 \textbf{FLIER} & 40 & 2h36min & \textbf{76.70} & \textbf{+14.72} \\
  \bottomrule 
  \end{tabular}
\end{table}
\begin{table}[tb]
  \caption{Flops and parameters of FLIER and CLIP.
  }
  \label{tab:3-1}
  \centering
  \begin{tabular}{@{}lcccc@{}}
    \toprule
    Models & Flops & Parameter \\
    \midrule
    CLIP& 17083.66 M & 82.46 M \\
    \midrule
    FLIER (image encoder) & 11271.12 M & 58.03 M \\
 FLIER (latent encoder) & 0.96 M & 0.77 M \\
  \bottomrule 
  \end{tabular}
\end{table}
\subsubsection{Performance on Other Benchmark Datasets.}
\begin{figure*}[t]
  \centering
  \begin{subfigure}{
    \includegraphics[width=3.2cm]{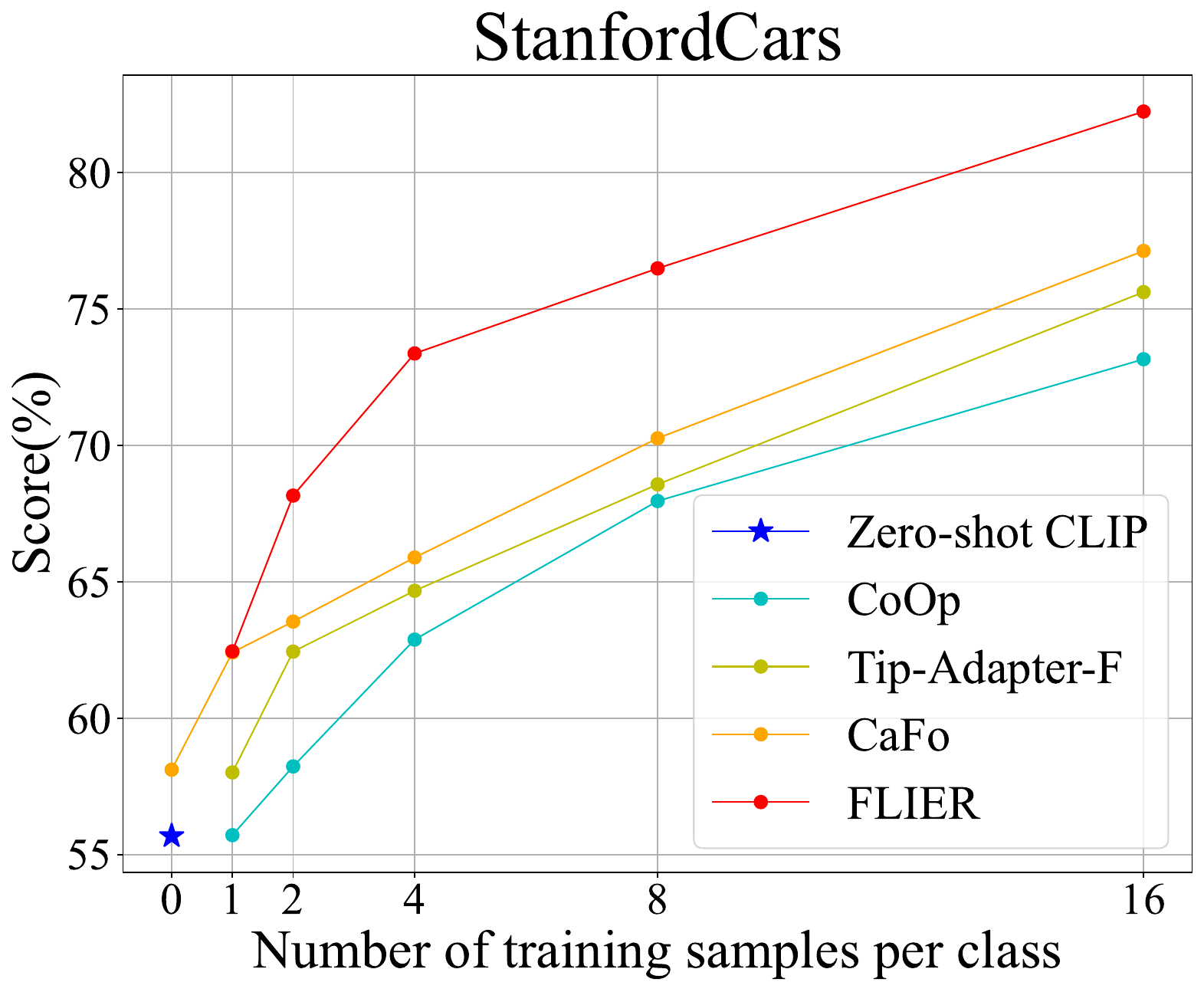}
    \label{fig:short-a}}
  \end{subfigure}
  \hfill
  \begin{subfigure}{
    \includegraphics[width=3.2cm]{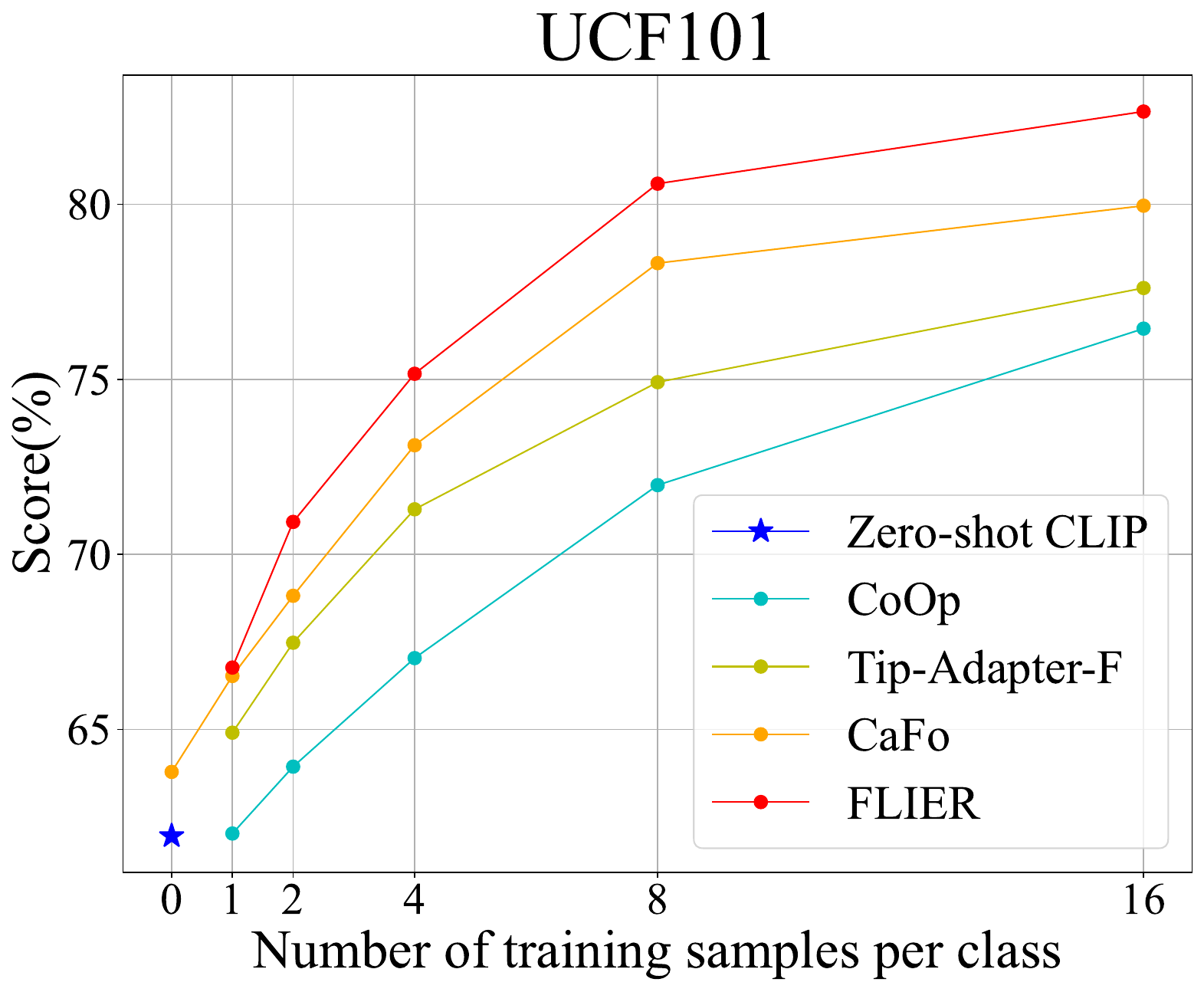}
    \label{fig:short-b}}
  \end{subfigure}
  \hfill
  \begin{subfigure}{
    \includegraphics[width=3.2cm]{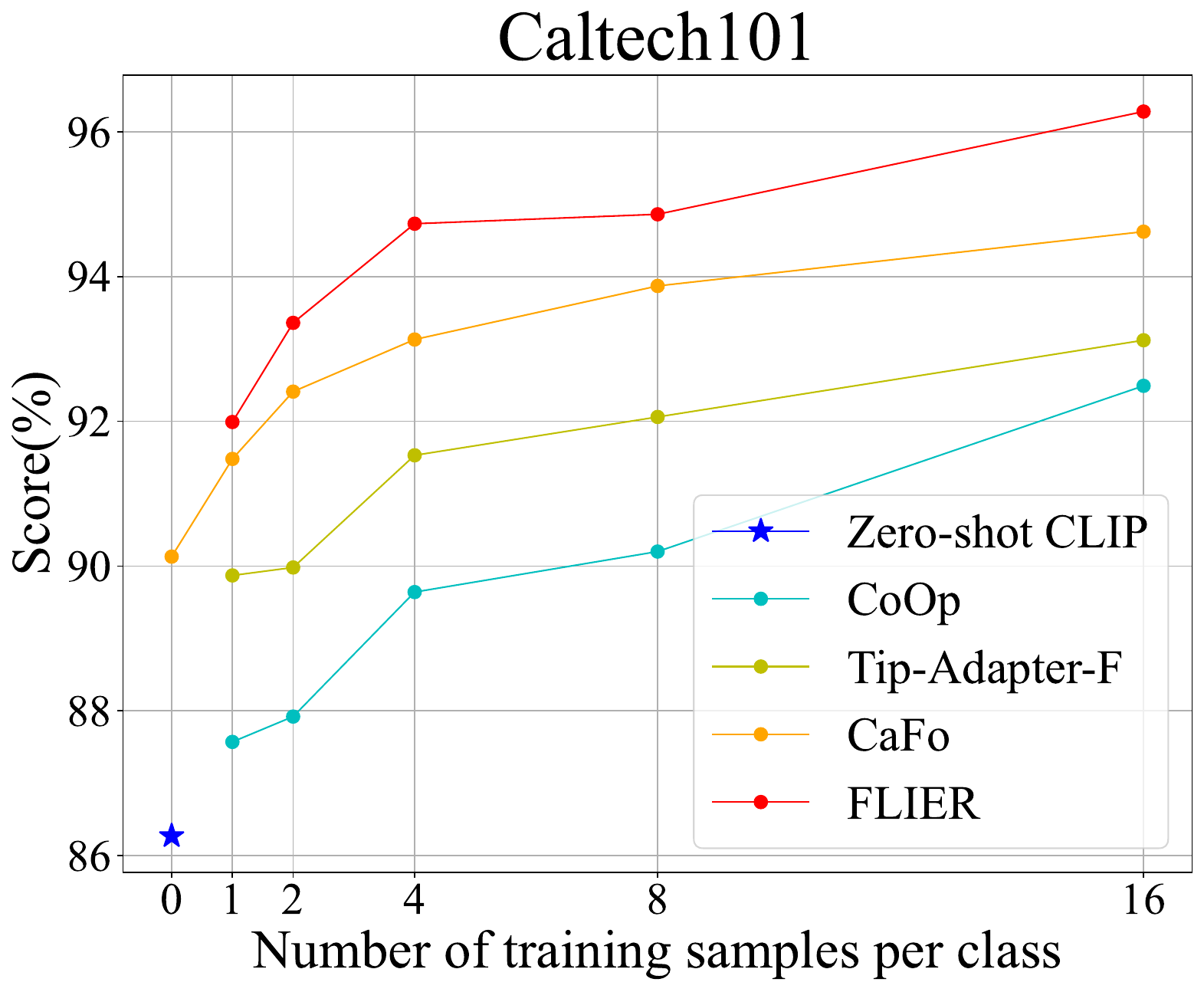}
    \label{fig:short-c}}
  \end{subfigure}
  \hfill
  \begin{subfigure}{
    \includegraphics[width=3.2cm]{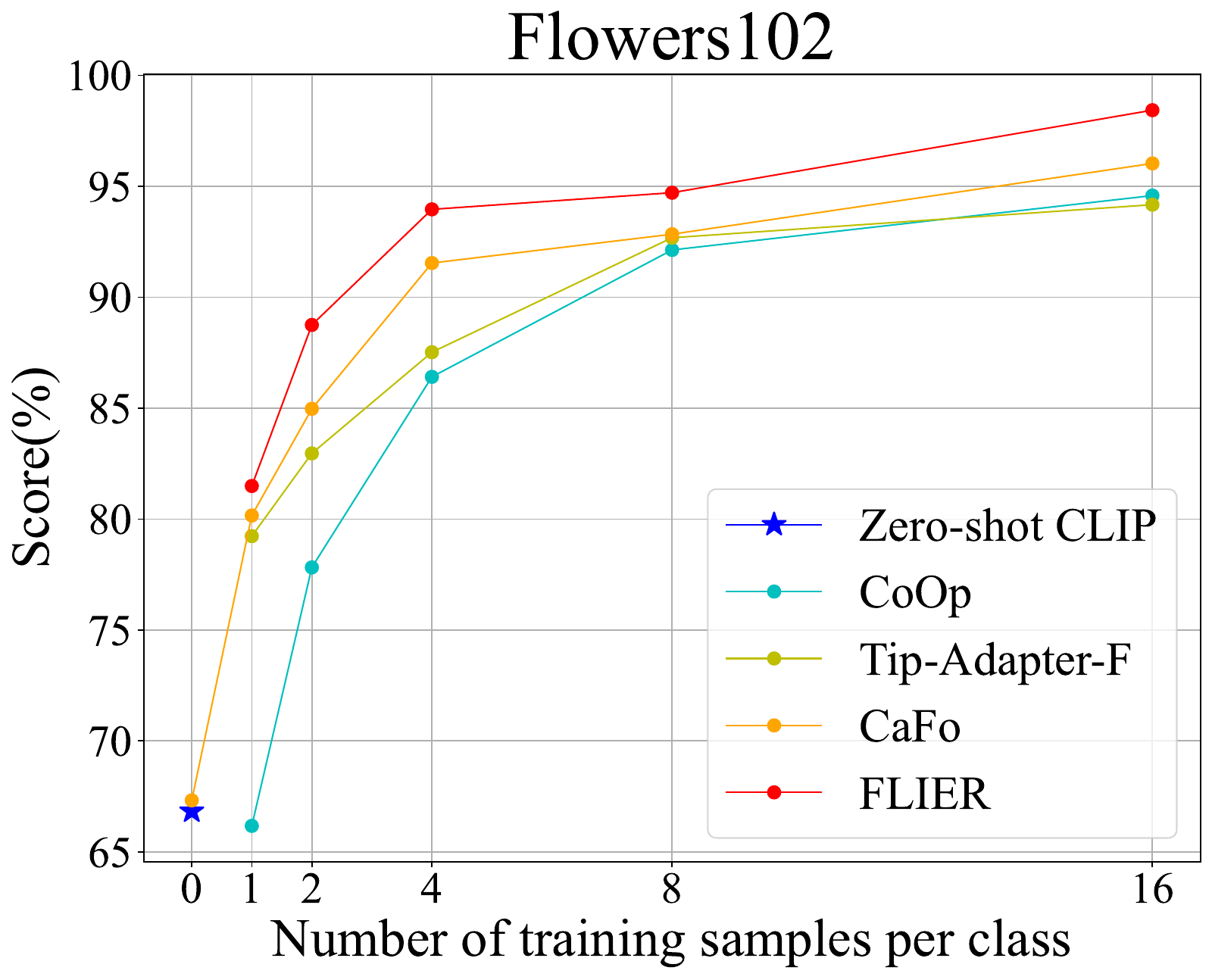}
    \label{fig:short-d}}
  \end{subfigure}
  \hfill
  \begin{subfigure}{
    \includegraphics[width=3.2cm]{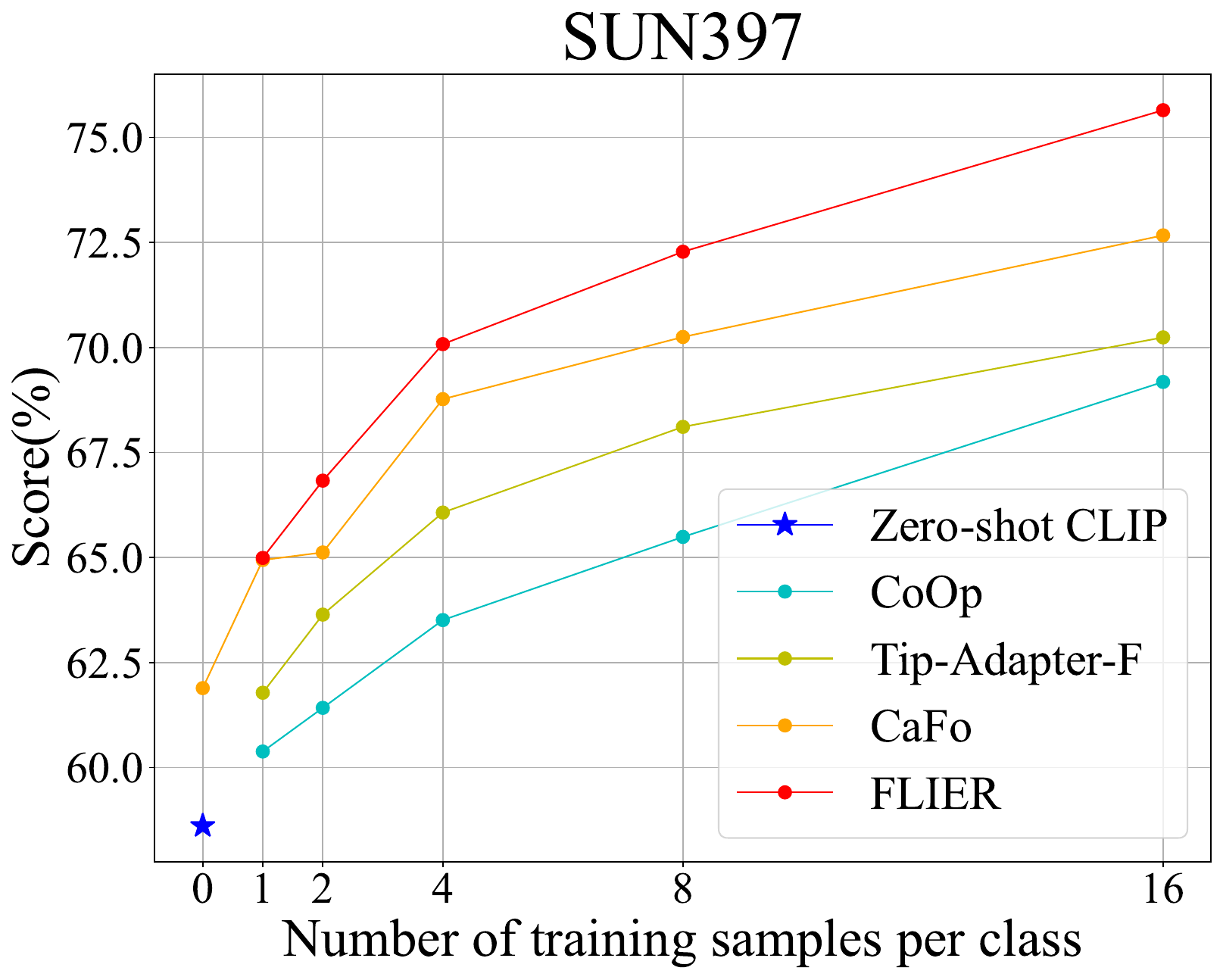}
    \label{fig:short-e}}
  \end{subfigure}
   \\
  \begin{subfigure}{
    \includegraphics[width=3.2cm]{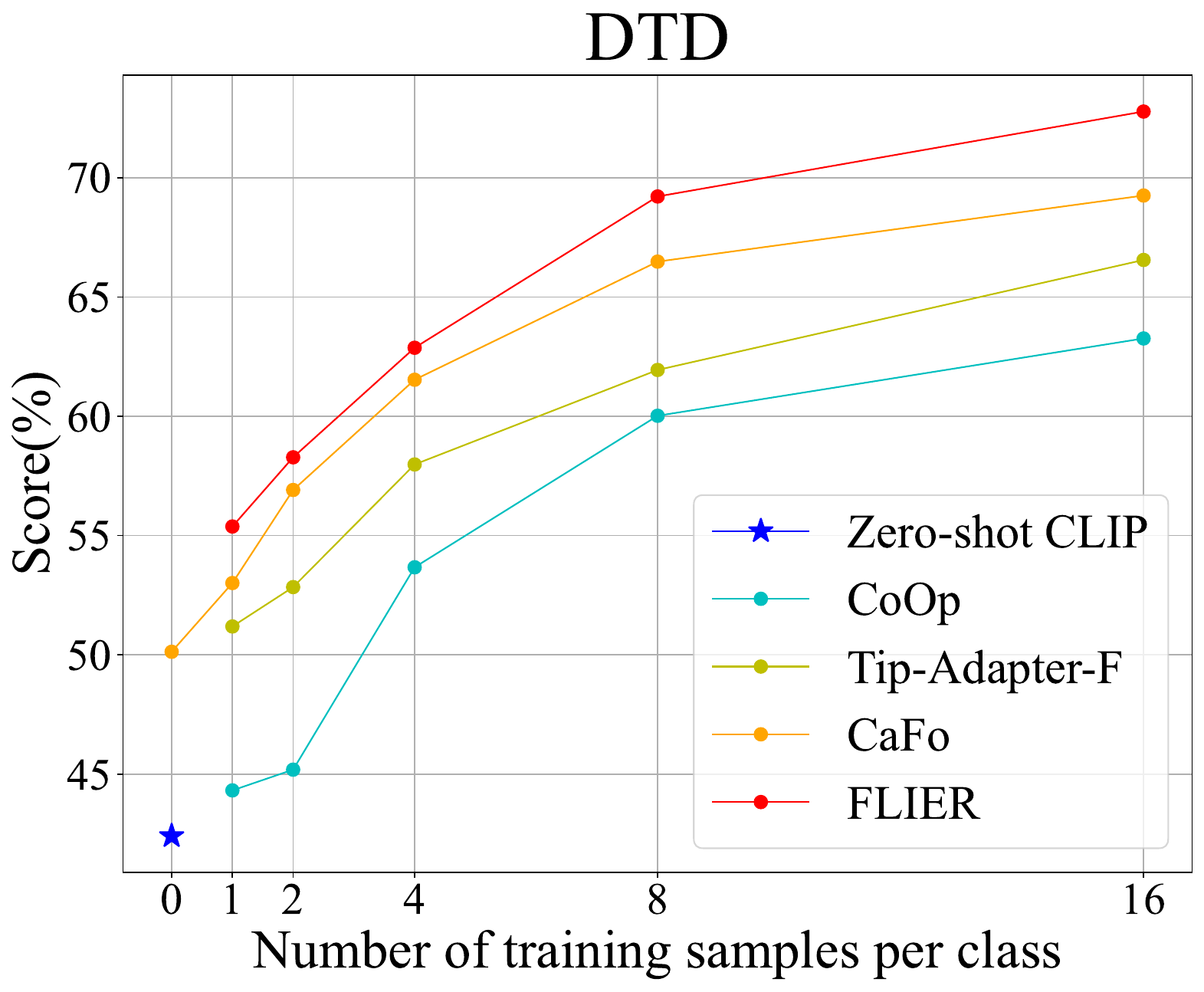}
    \label{fig:short-f}}
  \end{subfigure}
  \hfill
  \begin{subfigure}{
    \includegraphics[width=3.2cm]{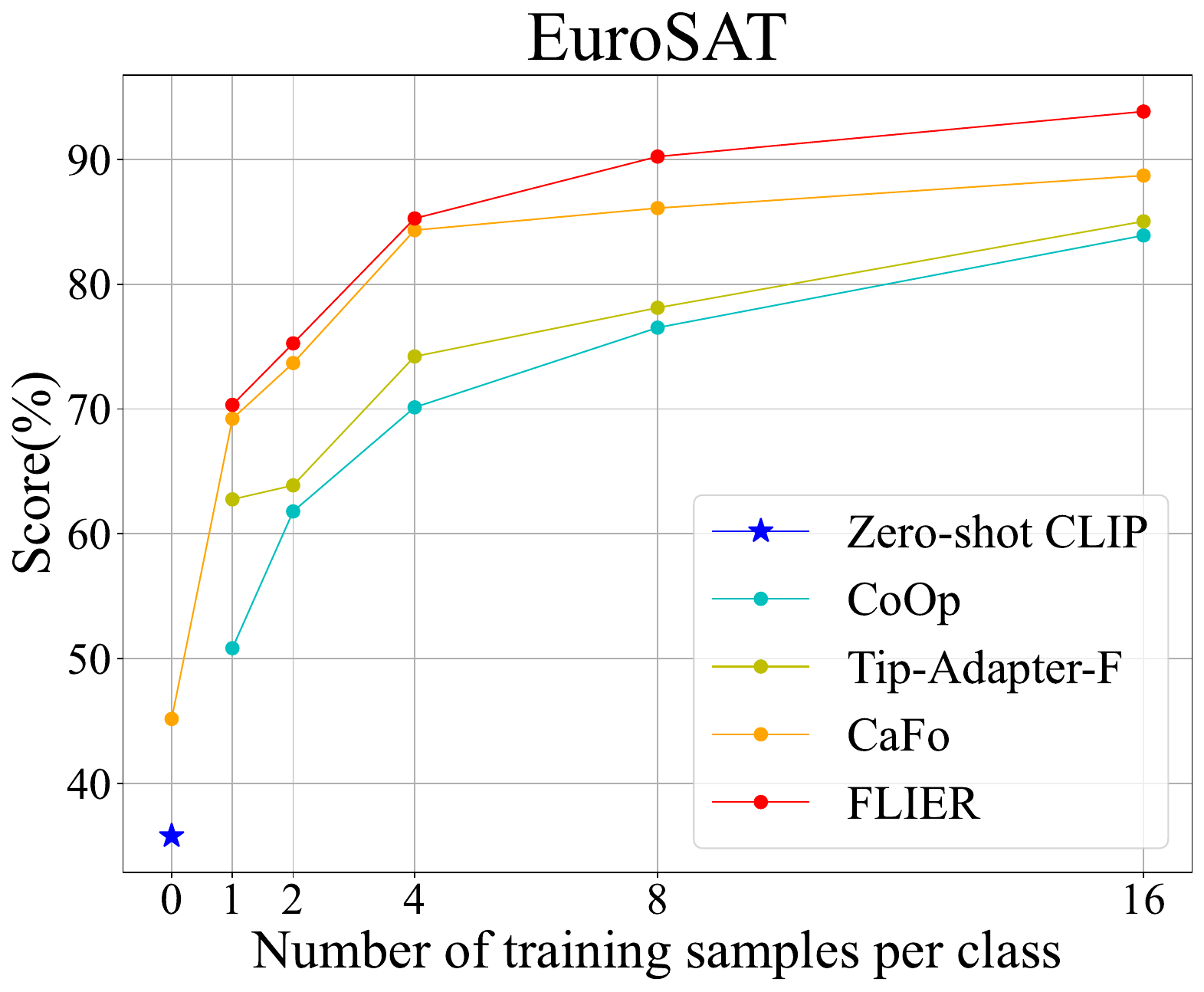}
    \label{fig:short-g}}
  \end{subfigure}
  \hfill
  \begin{subfigure}{
    \includegraphics[width=3.2cm]{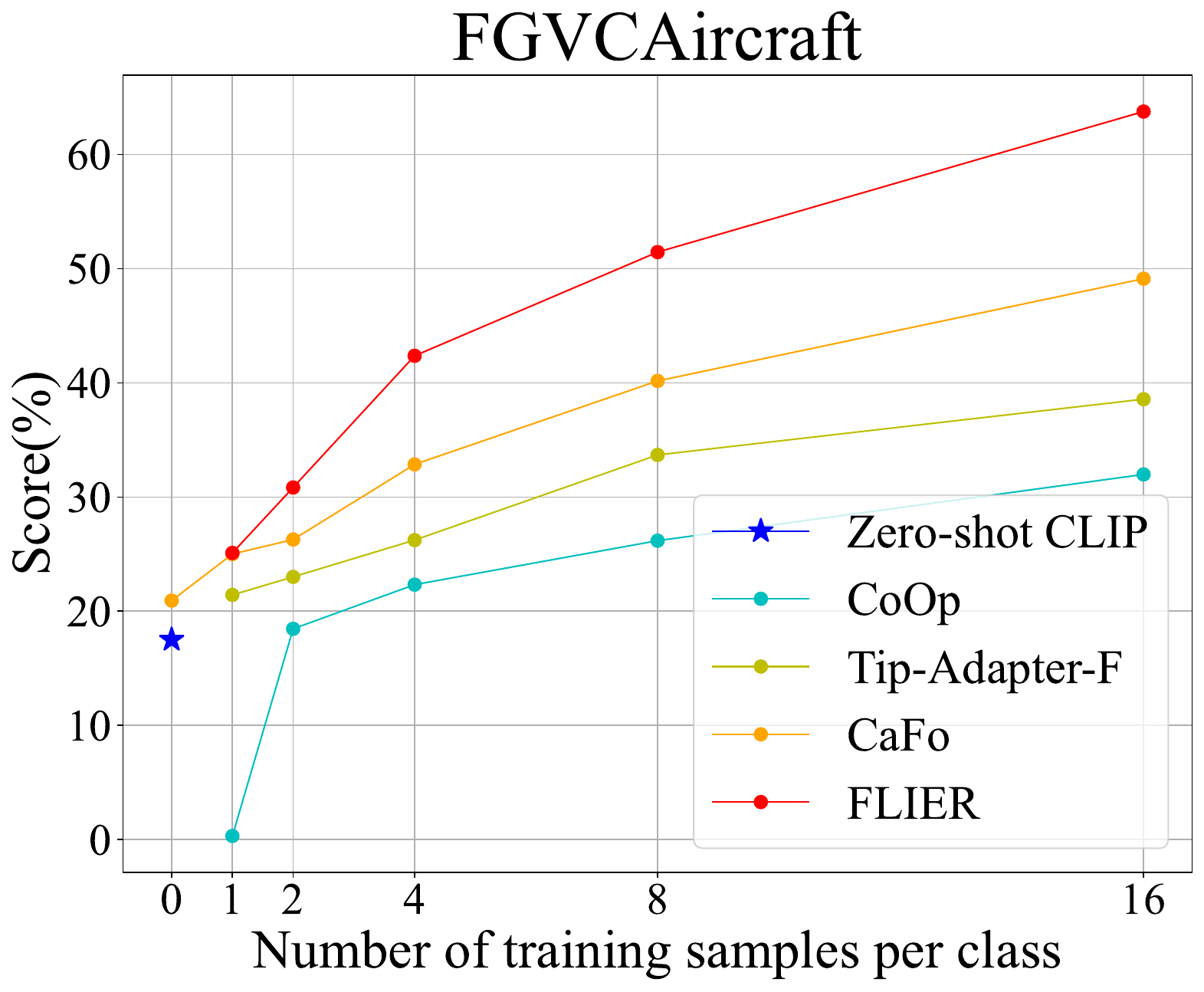}
    \label{fig:short-h}}
  \end{subfigure}
  \hfill
  \begin{subfigure}{
    \includegraphics[width=3.2cm]{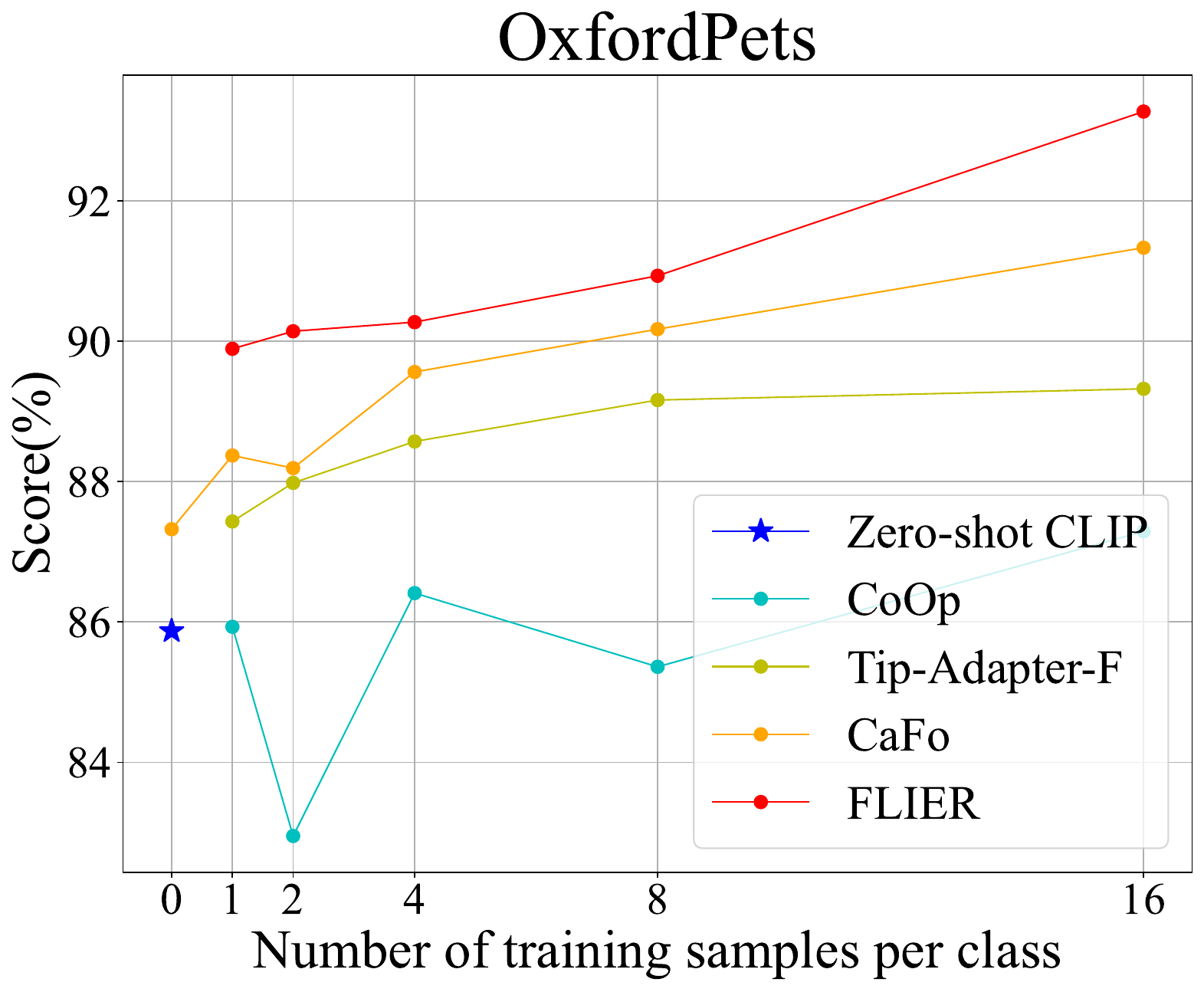}
    \label{fig:short-i}}
  \end{subfigure}
  \hfill
  \begin{subfigure}{
    \includegraphics[width=3.2cm]{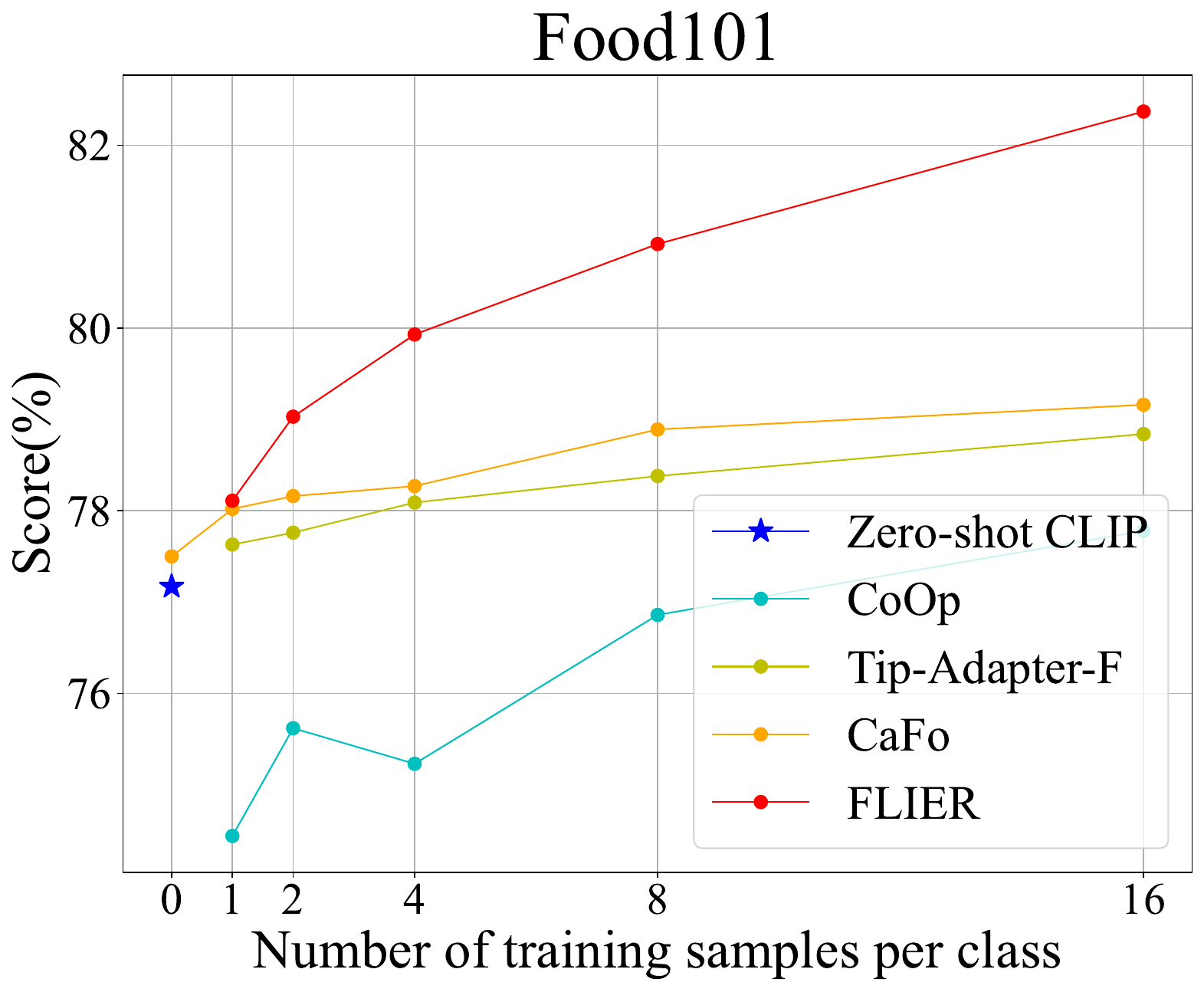}
    \label{fig:short-j}}
  \end{subfigure}
  \caption{Comparison of FLIER and other methods with backbone of RN50 on 10 datasets.}
  \label{fig:3}
\end{figure*}

To analyze the ability of FLIER comprehensively, we conduct experiments of FLIER on other 10 datasets and calculate the average results of 11 datasets with ImageNet. In \Cref{fig:2-2}, for the average accuracy on the 11 datasets, the accuracy of FLIER is higher than previous SOTA, CaFo with an increase of 0.46\%, 2.14\%, 2.67\%, 3.21\% and 3.87\% under 1-shot, 2-shot, 4-shot, 8-shot and 16-shot settings, respectively. The few-shot results on other 10 datasets are presented in \Cref{fig:3}. Overall, our FLIER outperforms all CLIP-based methods under all few-shot settings on 10 datasets. Of particular note is the performance of FLIER on FGVCAircraft, FLIER increases an accuracy of 9.51\%, 11.28\% and 14.64\% than SOTA's accuracy in 4-shot, 8-shot and 16-shot, respectively. Although FLIER shows a slight average improvement over the previous SOTA in the 1-shot setting, it outperforms the SOTA by 1.33\% on the Flowers102 dataset and 2.37\% on the DTD dataset. Another surprising result is that FLIER demonstrates superior performance in the 8-shot and 16-shot settings, averaging 3.21\% and 3.87\%, outperforming other methods by a large margin. FLIER also performs well in the 4-shot setting, with its accuracy surpassing that of all other baseline methods on Caltech101 and Food101 datasets, even outperforming their accuracy in the 16-shot setting.
\subsubsection{Data Domain Generalization.}
The ability to generalize to out-of-distribution data is important for deep learning models in real-world scenes and practical applications. We evaluate the domain generalization ability of FLIER by training on ImageNet and testing on ImageNet-V2\cite{recht2019imagenet} and ImageNet-Sketch\cite{hendrycks2021natural}, which contain the same categories with ImageNet. In \Cref{tab:4}, with latent encoder and prior knowledge from SD2, FLIER surpasses previous SOTA on ImageNet-V2 and ImageNet-Sketch by (1.05\%, 2.01\%) with RN50 backbone and (1.63\%, 0.99\%) with ViT-B/16 backbone. The results also demonstrate that FLIER with ViT backbone outperforms ResNet with a large margin.
\begin{table}[tb]
  \caption{Data Domain Generalization. We train the methods on ImageNet and test on ImageNet-V2/Sketch with RN50.}
  \label{tab:4}
  \centering
  \begin{tabular}{@{}lccccc@{}}
    \toprule
     \multirow{2}{*}{Datasets}  & \textbf{Source} & \multicolumn{2}{c}{\textbf{Target}} \\
      & ImageNet & -V2 & -Sketch \\
    \midrule
    Zero-shot CLIP & 60.33 & 53.27 & 35.44 \\
  CoOp & 62.95 & 54.58 & 31.40 \\
  CLIP-Adapter & 63.59 & 55.69 & 35.68 \\
  Tip-Adapter-F & 65.51 & 57.11 & 36.00 \\
  CaFo(RN50) & 68.79 & 57.99 & 39.43 \\
  CaFo(ViT-B/16) & 74.48 & 66.33 & 49.10 \\
  \midrule
  FLIER(RN50) &  70.03 & 59.04 & 41.44\\
  \textbf{FLIER(ViT-B/16)} &  \textbf{76.70} & \textbf{67.96} & \textbf{50.09}\\
  \bottomrule
  \end{tabular}
\end{table}
\subsection{Ablation studies}

\begin{table}[tb]
  \caption{Ablations on generated images on ImageNet with RN50. CLIP-AugData represents CLIP with generated data.
  }
  \label{tab:5}
  \centering
  \begin{tabular}{@{}lcccc@{}}
    \toprule
    Models & top-1-acc & top-5-acc & shot-1 & shot-2\\
    \midrule
    CLIP-finetune & 85.67 & 97.26 & 47.05 & 53.31 \\
    CLIP-AugData & 85.82 & 97.24& 47.30 & 54.72 \\
    \textbf{FLIER} & \textbf{87.08} & \textbf{98.65} & \textbf{61.51} & \textbf{64.43} \\
    \midrule
    Models & shot-4 & shot-8 & shot-16 & - \\
    CLIP-finetune & 61.71 & 66.85 & 68.35 & - \\
    CLIP-AugData & 62.22 & 66.89 & 68.42 & - \\
    \textbf{FLIER} & \textbf{65.98} & \textbf{68.86} & \textbf{70.03} & - \\
  \bottomrule
  \end{tabular}
\end{table}

\begin{table}[tb]
  \caption{Ablations on backbone of CLIP on ImageNet.
  }
  \label{tab:6}
  \centering
  \begin{tabular}{@{}lcccc@{}}
    \toprule
     \ (16-shot)  & RN50 & RN101 & ViT-B/32 & ViT-B/16 \\
  \midrule
  Zero-shot CLIP & 60.33 & 62.53 & 63.80 & 68.73\\
  CoOp & 62.95 & 66.60 & 66.85 & 71.92\\
  CLIP-Adapter & 63.59 & 65.39 & 66.19 & 71.13\\
  Tip-Adapter-F & 65.51 & 68.56 & 68.65 & 73.69\\
  CaFo & 68.79 & 70.86 & 70.82 & 74.48 \\
  \midrule
  \textbf{FLIER} &  \textbf{70.03} & \textbf{72.17} & \textbf{71.46}  & \textbf{76.70} \\
  \bottomrule
  \end{tabular}
\end{table}
\begin{table}[tb]
  \caption{Ablations on latent factor on 16-shot ImageNet.
  }
  \label{tab:7}
  \centering
  \begin{tabular}{@{}lccccc@{}}
    \toprule
     Latent factor & 0.1 & 0.3 & 0.5 & 0.7 & 0.9 \\
  \midrule
  FLIER(ViT-B/16) & 76.47 & 76.59 & \textbf{76.70}  & 76.51 & 76.23 \\
  \midrule
  \end{tabular}
\end{table}
\begin{table}[tb]
  \caption{Ablations on stages' order on 16-shot ImageNet.
  }
  \label{tab:8}
  \centering
  \begin{tabular}{@{}lccccc@{}}
    \toprule
     Shot & 1 & 2 & 4 & 8 & 16 \\
  \midrule
  FLIER(Joint 1st) & 65.21 & 67.09 & 68.71 & 72.72 & 76.68 \\
    FLIER & 65.22 & 67.11 & 68.63& 72.72 & 76.70 \\
  \bottomrule
  \end{tabular}
\end{table}
\subsubsection{Generated Images via Stable Diffusion 2.}
We conduct ablations to observe whether the generated images from SD2 help improve the performance of CLIP-finetune. We conduct the experiments both on the whole dataset and the few-shot setting. In \Cref{tab:5}, the performance of CLIP-finetune with generated images achieves a higher top-1 accuracy than CLIP-finetune, but a lower top-5 accuracy. In the few-shot results, CLIP-finetune with generated images show better performance than CLIP-finetune slightly. Overall, CLIP-finetune with generated images does not show obvious better performance than CLIP-finetune, which proves the effectiveness of another module of low-dimensional latent representations learned by SD2.
\subsubsection{Low-dimensional Latent Representations.}
After excluding the effectiveness of generated images, we conduct ablations on low-dimensional latent representations. Similar to the experiments in ablation of generated images, we perform ablations on whole ImageNet. In \Cref{tab:5}, FLIER exceeds CLIP-finetune and CLIP-AugData obviously with an increase of 1.41\%, 1.26\% in top-1 accuracy and 1.39\%, 1.41\% in top-5 accuracy. This demonstrates the effectiveness of the general vision representation learning in FLIER. In addition, FLIER is a strong few-shot learner than CLIP-finetune, outperforming that by an increase of 17.46\%, 13.10\%, 4.27\%, 2.01\% and 1.68\% in accuracy under 1-shot, 2-shot, 4-shot, 8-shot and 16-shot, respectively. By two ablation studies, we conclude that additional data is not the main reason for FLIER’s superior performance, instead, the low-dimensional latent representations work in FLIER.

\subsubsection{CLIP’s Visual Encoders.}
In FLIER, we conducted experiments using various visual encoders of CLIP to demonstrate the versatility of FLIER across different backbones. As shown in \Cref{tab:6}, the performances of FLIER with ViT-B/16 are higher than other visual backbones with a large margin. In addition, FLIER outperforms all other baselines across different backbones, indicating the effectiveness of FLIER with different network architectures.

\subsubsection{Latent Factor.}
The latent factor in FLIER is to control the effect that the latent encoder contributes to the whole model. We explore the model's performance with different latent factors from 0.1 to 0.9 ($\alpha=0.1, 0.3, 0.5, 0.7, 0.9$) on ImageNet under 16-shot setting in the backbone of ViT-B/16. From \Cref{tab:7}, we observe that neither too large nor too small latent factors contribute to better few-shot performance. Since CLIP's image encoder is trained twice with both generated images and training images, a larger latent factor might not lead to a thorough fine-tuning of the image encoder, which affects the performance. On the contrary, when the latent factor is too small, the contributions from the latent encoder are significantly reduced, causing the model to behave close to the original CLIP-finetune. When the latent factor is 0.5, the model's performance in experiments is relatively higher than that in other experiments, demonstrating that FLIER benefits the most from the relative average contribution from the latent encoder and image encoder.

\subsubsection{Order of two stages.}
We swap the order of two stages to explore whether the stages' order would influence the performance of FLIER. We conduct the experiment on FLIER with the backbone of ViT-B/16 on the setting of 16-shot on ImageNet. FLIER(Joint 1st) represents that we train the joint training phrase first and train the image encoder later in one FLIER's training stage. From \Cref{tab:8}, there are no obvious differences between two different orders, indicating the order does not affect the performance of FLIER.

\subsection{Images-only architecture with an image-to-image generation module}
Taking into account the potential benefits of textual information within the FLIER framework, we replace the text-to-image generation module of SD2 with an image-to-image generation module, setting the corresponding prompt to an empty string. In this framework, the latent representation generated by the image-to-image module is aligned with the shape of CLIP's image encoder through a trainable deconvolution layer. The image-to-image sampler, deconvolution layer, and image encoder were trained concurrently. Experimental results indicate that, after modifying the framework to image-to-image generation, the model fails to converge and yields very low accuracy on the test set. We hypothesize that the upsampling performed by the deconvolution layer within the pipeline, which inputs the latent representation into the image encoder, significantly degrades the quality of the latent representation. Also, the ViT with a fixed patch size and pre-trained weights produces a large number of redundant and non-informative tokens during image tokenization, leading to poor training performance. These findings support the use of diffusion models  of FLIER over others.


\section{Limitations}
While we believe FLIER to be a robust few-shot learner, we have not conducted experiments on professional application datasets. Therefore, the performance of FLIER in real-world scene applications remains uncertain. In the future, we aim to conduct additional experiments to evaluate FLIER's practical applicability. In addition, due to the characteristics of fine-tuning and few-shot setting, FLIER's performance in 1-shot setting is not as strong as more shots setting.

\section{Conclusion}
In this paper, we explored the feasibility of low-dimensional latent representations generated from image synthesis tasks for image recognition and few-shot learning. Leveraging GPT-3 and SD2, we obtained latent representations, which were then embedded into a vision-language model, CLIP, through the latent encoder architecture of FLIER. To further investigate the auxiliary effects of latent representations on FLIER, we introduced latent factors to quantitatively explore the performance of FLIER with different contributions of latent representations. In the few-shot experiments across 11 datasets, FLIER has demonstrated SOTA performance for most tasks. In the future, we plan to investigate the possibility of constructing an independent backbone, possibly based on ViT, with latent representations as the core module. This backbone could potentially be versatile and applicable across various tasks.

We declare that we will release source code upon acceptance of the paper.

\bibliography{aaai25}

\end{document}